\pdfoutput=1

\documentclass[11pt]{article}

\usepackage{emnlp2021}
\usepackage{times}
\usepackage{latexsym}
\usepackage{booktabs}

\usepackage{selectp}

\usepackage{microtype}

\usepackage{amsmath} 
\usepackage{amsfonts} 
\usepackage[T1]{fontenc} 
\usepackage{array, multirow} 
\usepackage{graphicx} 
\usepackage{nicefrac}

\usepackage{etoolbox,siunitx}
\robustify\bfseries

\usepackage[utf8]{inputenc}

\DeclareMathOperator{\softmax}{softmax}
\DeclareMathOperator{\argmax}{argmax}

\title{Structured Context and High-Coverage Grammar for Conversational Question Answering over Knowledge Graphs}

\author{Pierre Marion$\thanks{\; Work done while an intern at Google Research.}$ \\
  Sorbonne Université, CNRS \\
  Laboratoire de Probabilités, Statistique et Modélisation, LPSM \\
  F-75005, Paris, France \\
  \texttt{pierre.marion@sorbonne-universite.fr} 
  \AND
  Pawe{\l} Krzysztof Nowak \and Francesco Piccinno \\
  Google Research \\
  \texttt{\{pawelnow,piccinno\}@google.com}}

\begin{document}

\maketitle
\sisetup{detect-weight=true,detect-inline-weight=math}

\begin{abstract}
We tackle the problem of weakly-supervised conversational Question Answering over large Knowledge Graphs using a neural semantic parsing approach. 
We introduce a new Logical Form (LF) grammar that can model a wide range of queries on the graph while remaining sufficiently simple to generate supervision data efficiently. 
Our Transformer-based model takes a JSON-like structure as input, allowing us to easily incorporate both Knowledge Graph and conversational contexts. This structured input is transformed to lists of embeddings and then fed to standard attention layers. 
We validate our approach, both in terms of grammar coverage and LF execution accuracy, on two publicly available datasets, CSQA and ConvQuestions, both grounded in Wikidata. On CSQA, our approach increases the coverage from $80\%$ to $96.2\%$, and the LF execution accuracy from $70.6\%$ to $75.6\%$, with respect to previous state-of-the-art results. On ConvQuestions, we achieve competitive results with respect to the state-of-the-art.
\end{abstract}

\section{Introduction}
Graphs are a common abstraction of real-world data. Large-scale knowledge bases can be represented as directed labeled graphs, where entities correspond to nodes and subject-predicate-object triplets are encoded by labeled edges. These so-called \textit{Knowledge Graphs} (KGs) are used both in open knowledge projects (YAGO, Wikidata) and in the industry (Yahoo, Google, Microsoft, etc.). A prominent task on KGs is \textit{factual conversational Question Answering} (Conversational KG-QA) and it has spurred interest recently, in particular due to the development of AI-driven personal assistants. 

The Conversational KG-QA task involves difficulties of different nature: entity disambiguation, long tails of predicates \cite{sahaComplexSequentialQuestion2018}, conversational nature of the interaction. The topology of the underlying graph is also problematic. Not only can KGs be huge (up to several billion entities), they also exhibit hub entities with a large number of neighbors.

A recent prominent approach has been to cast the problem as neural \textit{semantic parsing} \cite{jia-liang-2016-data, dong-lapata-2016-language, dong-lapata-2018-coarse, shen-etal-2019-multi}. In this setting, a semantic parsing model learns to map a natural language question to a \textit{logical form} (LF), \textit{i.e.} a tree of operators over the KG. These operators belong to some grammar, either standard like SPARQL or ad-hoc. The logical form is then \textit{evaluated} over the KG to produce the candidate answer. 
In the \textit{weak supervision} training setup, the true logical form is not available, but only the answer utterance is (as well as annotated entities in some cases, see Section \ref{subsec:nel}). Hence the training data is not given but it is instead generated, in the format of \textit{(question, logical form)} pairs. We refer to this data as \textit{silver data} or \textit{silver LFs}, as opposed to unknown gold ground truth. 

However, this approach has two main issues. First, the silver data generation step is a complex and often resource-intensive task. The standard procedure employs a Breadth-First Search (BFS) exploration \cite{guoDialogtoActionConversationalQuestion2018, shen-etal-2019-multi}, but this simple strategy is prone to failure, especially when naively implemented, for questions that are mapped to nested LFs. This reduces the \textit{coverage}, i.e. the percentage of training questions associated to a Logical Form.
\citet{shenEffectiveSearchLogical2019}~proposes to add a neural component for picking the best operator, in order to reduce the computational cost of this task, however complicating the model. \citet{cao-etal-2020-unsupervised-dual} proposes a two-step semantic parser: the question is first paraphrased into a ``canonical utterance'', which is then mapped to a LF. This approach simplifies the LF generation by separating it from the language understanding task.

Second, most of the semantic parsing models do not leverage much of the underlying KG structure to predict the LF, as in \citet{dong-lapata-2016-language,guoDialogtoActionConversationalQuestion2018}. Yet, this contextual graph information is rich \cite{tongLeveragingDomainContext2019}, and graph-based models leveraging this information yield promising results for KG-QA tasks \cite{vaku-etal-2019CIKM, christmannLookYouHop2019}.
However these alternative approaches to semantic parsing, that rely on node classification, have their inherent limitations, as they handle less naturally certain queries (see Appendix \ref{subsec:comparison-node-classif}) and their output is less interpretable. This motivates the desire for semantic parsing models that can make use of the KG context.

\paragraph{Approach and contributions}
We design a new grammar, which can model a large range of queries on the KG, yet is simple enough for BFS to work well. 
We obtain a high coverage on two KG-QA datasets. On CSQA \cite{sahaComplexSequentialQuestion2018}, we achieve a coverage of $96\%$, a $16\%$ improvement over the baseline \cite{shenEffectiveSearchLogical2019}. On ConvQuestions \cite{christmannLookYouHop2019}, a dataset with a large variety of queries, we reach a coverage of $86\%$.

To leverage the rich information contained in the underlying KG, we propose a semantic parsing model that uses the KG contextual data in addition to the utterances. Different options could be considered for the KG context, \textit{e.g.} lists of relevant entities, annotated with metadata or pre-trained entity embeddings that are graph-aware \cite{zhangGraphBertOnlyAttention2020}. The problem is that this information does not come as unstructured textual data, which is common for language models, but is structured.

To enable the use of context together with a strong language model, we propose the Object-Aware Transformer (OAT) model, which can take as input structured data in a JSON-like format. The model then transforms the structured input into embeddings, before feeding them into standard Transformer layers.
With this approach, we improve the overall execution accuracy on CSQA by $5.0\%$ compared to a strong baseline \cite{shen-etal-2019-multi}. On ConvQuestions, we improve the precision by $4.7\%$ compared to \citet{christmannLookYouHop2019}. 

\section{Related work}
\label{sec:related-work}

\paragraph{Neural semantic parsing} Our work falls within the neural semantic parsing approaches for Knowledge-Based QA~\cite{dong-lapata-2016-language, liang-etal-2017-neural, dong-lapata-2018-coarse,guo-etal-2019-towards, hwangComprehensiveExplorationWikiSQL2019}.
The more specific task of conversational KG-QA has been the focus of recent work. 
\citet{guoDialogtoActionConversationalQuestion2018}~introduces D2A, a neural symbolic model with memory augmentation. This model has been extended by S2A+MAML \cite{guo-etal-2019-coupling} with a meta-learning strategy to account for context, and by D2A+ES \cite{shenEffectiveSearchLogical2019} with a neural component to improve BFS. \citet{saha-etal-2019-complex} proposes a Reinforcement Learning model to benefit from denser supervision signals. Finally, \citet{shen-etal-2019-multi} introduces MaSP, a multi-task model that performs both entity linking and semantic parsing, with the hope of reducing erroneous entity linking. Recently, \citet{plepi2021context} extended the latter in CARTON. They first predict the LF using a Transformer architecture, then specify the KG items using pointer networks.

\paragraph{Learning on Knowledge Graphs} Classical graph learning techniques can be applied to the specific case of KGs. In \textsc{Convex} \citep{christmannLookYouHop2019}, at each turn, a subgraph is expanded by matching the utterance with neighboring entities. Then a candidate answer is found by a node classifier. Other methods include unsupervised message passing \cite{vaku-etal-2019CIKM}. However, these approaches lack strong NLP components. Other directions include learning differentiable operators over a KG \cite{cohenNeuralQueryLanguage2019}, or applying Graph Neural Networks \cite{kipfSemiSupervisedClassificationGraph2017, hamiltonInductiveRepresentationLearning2017} (GNNs) to the KG, which has been done for entity classification and link prediction tasks \cite{schlichtkrullModelingRelationalData2018}. GNNs have also been used to model relationships between utterances and entities \cite{shaw-etal-2019-generating}. 

\paragraph{Structured Input for neural models} Our approach of using JSON-like input falls in the line of computing neural embeddings out of structured inputs. 
\citet{tai-etal-2015-improved} introduced Tree-LSTM for computing tree embeddings bottom-up. It has then been applied for many tasks, including computer program translation \cite{chenTreetotreeNeuralNetworks2018}, semantic tree structure learning (such as JSON or XML) \cite{woofFrameworkEndtoEndLearning2020} and supervised KG-QA tasks \cite{tongLeveragingDomainContext2019, zafarDeepQueryRanking2019, athreyaTemplatebasedQuestionAnswering2020}. In the latter context, Tree-LSTM is used to model the syntactic structure of the question. Other related approaches include tree transformer \cite{harerTreeTransformerTransformerBasedMethod2019} and tree attention \cite{ahmedImprovingTreeLSTMTree2019}. Syntactic structures were also modeled as graphs \cite{xu-etal-2018-exploiting, li-etal-2020-graph-tree}. Specific positional embeddings can also be used to encode structures \cite{shivNovelPositionalEncodings2019, herzig-etal-2020-tapas}.

\section{A grammar for KG exploration}

\begin{table*}[t]
\centering
\begin{tabular}{m{0.1\linewidth}m{0.22\linewidth}m{0.14\linewidth}m{0.43\linewidth}}
\toprule
\textbf{Category} &  \textbf{Name} & \textbf{Signature} & \textbf{Description} \\
\midrule
\multirow[c]{5}{\linewidth}{Graph \\ operators} & \texttt{follow\_property} & (SE, P) $\rightarrow$ SE & 
Returns the \textbf{entities} which are linked by property P \textbf{to} at least one element of SE. \\\cline{2-4}
& \texttt{follow\_backward} & (SE, P) $\rightarrow$ SE & 
Returns the \textbf{entities} which are linked by property P \textbf{from} at least one element of SE. \\\cline{2-4}
& \texttt{get\_value} & (SE, P) $\rightarrow$ SV & 
Returns the \textbf{values} which are linked by property P to at least one element of SE. \\
\hline
\multirow[c]{5}{\linewidth}{Numerical operators} & \texttt{max},\texttt{ min} & SV $\rightarrow$ SV & 
Returns the max (resp. min) value from SV. \\\cline{2-4}
& \texttt{greater\_than}, \texttt{equals}, \texttt{lesser\_than} & (SV, V) $\rightarrow$ SV & 
Filters SV to keep values strictly greater than (resp. equal to, strictly lesser than) V. \\\cline{2-4}
& \texttt{cardinality} & SE $\rightarrow$ V & 
Returns the cardinality of SE. \\
\hline
\multirow[c]{4}{\linewidth}{Set operators} & \texttt{is\_in} & (a: SE, b: SE)  $\rightarrow$ SV &
Returns a boolean set: for each entity in a, the mask equals True if the entity is in b. \\\cline{2-4}
& \texttt{get\_first} & SE  $\rightarrow$ SE & 
Returns the 
first entity from SE. \\\cline{2-4}
& \texttt{union}, \texttt{intersect}, \texttt{difference} & (SE,SE)  $\rightarrow$ SE & 
Returns the union (resp. intersection, difference) of input sets. \\
\hline
\multirow[c]{2}{\linewidth}{Class operators} & \texttt{members} & SC  $\rightarrow$ SE &
Returns the members of classes in SC. \\\cline{2-4}
& \texttt{keep} & (SE,SC)  $\rightarrow$ SE &
Filters SE to keep the members of SC. \\
\hline
\multirow[c]{6}{\linewidth}{Meta-operators} & \texttt{for\_each} & SE  $\rightarrow$ SE &
Initializes a parallel computation over all entities in the input set.  \\\cline{2-4}
& \texttt{arg} & SV  $\rightarrow$ SE \textit{or} SE $\rightarrow$ SE&
Ends a parallel computation by returning all entities that gave a non-empty result.  \\\cline{2-4}
& \texttt{argmax}, \texttt{ argmin} & SV  $\rightarrow$ SE &
Ends a parallel computation by returning all entities that gave the max (resp. min) value.  \\
\bottomrule
\end{tabular}
\caption{\label{list-operators} List of operators in our grammar. Their variables can be entities (E), classes (C), values (V), ordered sets of such elements (resp. SE, SC and SV), or properties (P).}
\end{table*}

Several previous KG-QA works were based on the grammar from D2A \citep{guoDialogtoActionConversationalQuestion2018}. We also take inspiration from their grammar, but redesign it to model a wider range of queries. By defining more generic operators, we achieve this without increasing the number of operators nor the average depth of LFs. 
Section \ref{subsec:main-paper-comparison-d2a} presents a comparison.

\subsection{Definitions}

An \textit{entity} (e.g. \texttt{Marie Curie}) is a node in the KG. Two entities can be related through a directed labeled edge called a \textit{property} (e.g. \texttt{award received}). A property can also relate an entity to a \textit{value}, which can be a date, a boolean, a quantity or a string. Entities and properties have several attributes, prominently a name and an integer ID. The \texttt{membership} property is treated separately; it relates a member entity (e.g. \texttt{Marie Curie}) to a \textit{class} entity (e.g. \texttt{human being}).  

The \textit{objects} we will consider in the following are entities, properties, classes, and values. The grammar consists of a list of \textit{operators} that take objects or sets of objects as arguments. A Logical Form is a binary expression tree of operators.

In several places, we perform \textit{Named Entity Linking (NEL)}, i.e. mapping an utterance to a list of KG objects.
Section \ref{subsec:nel} details how this is done.

Table~\ref{list-operators} lists the operators we use, grouped in five categories. Most of them are straightforward, except meta-operators, which we explain next.

\subsection{Meta-operators}

Meta-operators are useful for questions such as: \emph{``Which musical instrument is played by the maximum number of persons?''}. To answer this question, we first compute the set of all musical instruments in the KG. For each entity in this set, we then follow the property \texttt{played by}, producing a set of people who play that instrument. Finally, we compute the max cardinality of all these sets and return the associated instrument.

The corresponding LF is the following:

{
\small
\noindent \texttt{argmax( \\
\indent cardinality( \\
\indent \indent follow\_property( \\
\indent \indent \indent for\_each( \\
\indent \indent \indent \indent members(\textit{musical instrument}), \\
\indent \indent \indent ), \\
\indent \indent \indent \textit{played by})))
}
}

\texttt{for\_each} creates a parallel computation over each entity in its argument, which can be terminated by three operators (\texttt{arg}, \texttt{argmax} and \texttt{argmin}). We refer to Appendix \ref{subsec:meta-operators} for details.

\subsection{Silver LF generation}

To generate silver LFs, we explore the space of LFs with a BFS strategy, similarly to \citet{guoDialogtoActionConversationalQuestion2018, shen-etal-2019-multi}. 
More precisely, to initialize the exploration, we perform NEL to find relevant entities, values and classes that appear in the question.
LFs of depth $0$ simply return an annotated object. 
Then, assume that LFs of depth less or equal to $n$ have been generated and we want to generate those of depth $n+1$. We loop through all possible operators; for each operator, we choose each of its arguments among the already-generated LFs. 
This algorithm brings two challenges, as highlighted in \citet{shenEffectiveSearchLogical2019}: computational cost and spurious LFs. We refer to  Appendix \ref{appendix:silver-lf} for implementation details that mitigate these difficulties.

\subsection{Comparison with D2A \citep{guoDialogtoActionConversationalQuestion2018}}
\label{subsec:main-paper-comparison-d2a}

Section \ref{subsec:results} shows that our grammar achieves higher coverage with a similar average LF depth. A more thorough quantitative comparison is delicate, as it would require reimplementing D2A within our framework, which is beyond the scope of this paper. On a qualitative basis, we use more elementary types: in addition to theirs, we introduce set of classes, strings and set of values (which can be strings, numerals or booleans). We use eight less operators than D2A; among our operators, six are in common (\texttt{follow\_property}, \texttt{follow\_backward}, \texttt{cardinality}, \texttt{union}, \texttt{intersect}, \texttt{difference}), four are modified (\texttt{keep}, \texttt{is\_in}, \texttt{argmax}, \texttt{argmin}), and the other ten are new. New intents that can be modeled include numerical reasoning (e.g. \textit{What actor plays the younger child?}), temporal reasoning (e.g. \textit{What is the number of seasons until 2018?}), ordinal reasoning (e.g. \textit{What was the first episode date?}), textual form reasoning (e.g. \textit{What was Elvis Presley given name?}). We refer to Appendix \ref{subsec:comparison-d2a} for more details and comparison methodology.

\section{Model}

\begin{figure*}
    \centering
    \includegraphics[width=\textwidth]{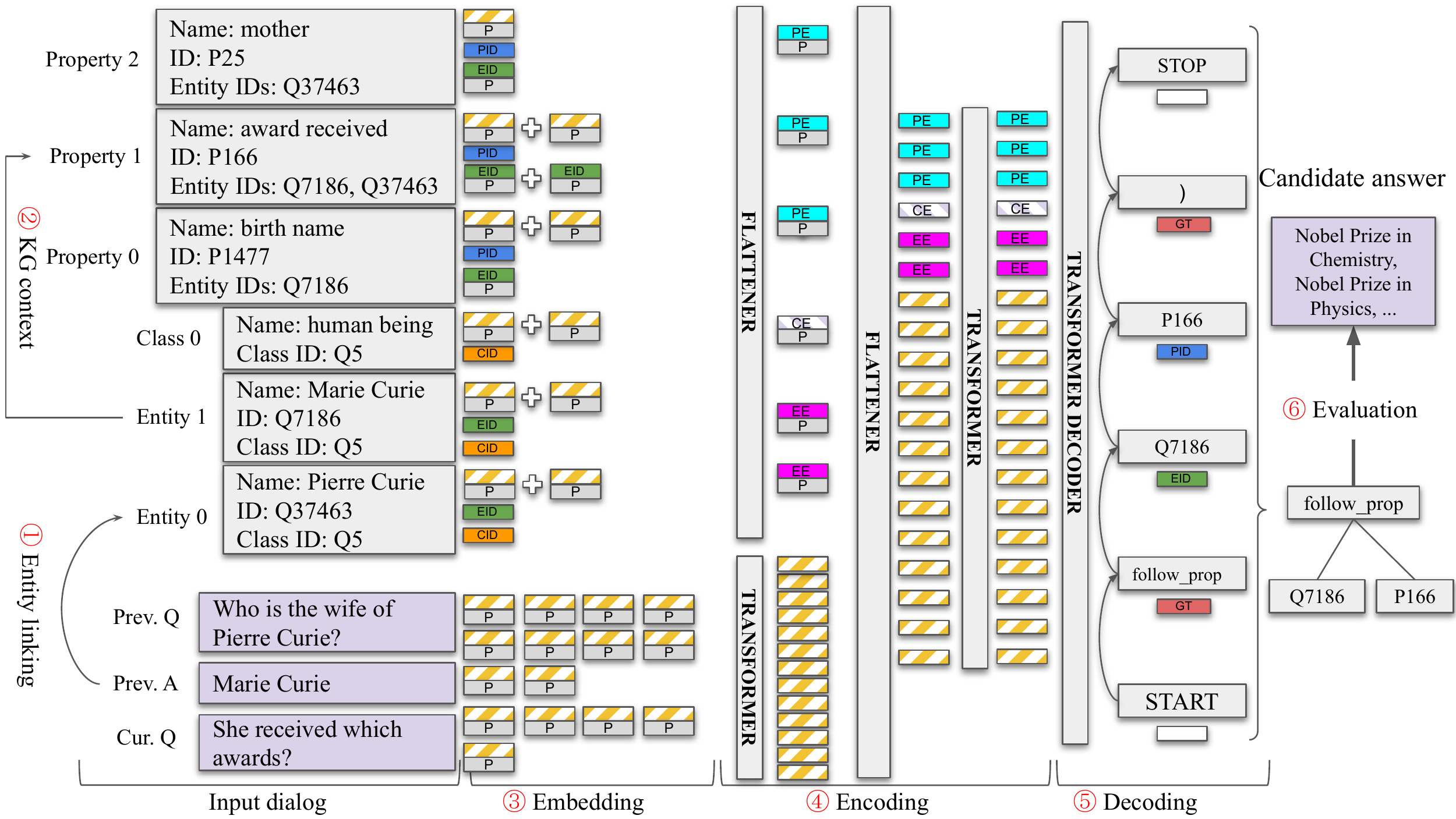}
    \caption{Architecture of the proposed model. The initial field embeddings are Positional (P), Property ID (PID), Entity ID (EID), and Class ID (CID). After the first Flattener layer, we obtain Property Embeddings (PE), Class Embeddings (CE), Entity Embeddings (EE). There are also Grammar Token (GT) embeddings in the output. Note that the entity IDs are actually randomized (not shown here).} 
    \label{fig:model}
\end{figure*}

\subsection{Overview}

The model, called Object-Aware Transformer (OAT), is a Transformer-based \cite{vaswaniAttentionAllYou2017a} auto-regressive neural semantic parser.
As illustrated in Figure~\ref{fig:model}, the model has several steps.

The first step consists of retrieving relevant objects, annotated with metadata, that might appear in the resulting LF. This step is performed using NEL on the utterances and KG lookups to retrieve the graph context information. At this point, the input is composed of lists of \textit{objects} with their \textit{fields}. 
After embedding each field value in a vector space, we perform successive layers of input flattening and Transformer encoding. The \textit{Flattener} layer is useful to transform the structured input into a list of embeddings. 
Then a decoder Transformer layer produces a linearized LF, i.e. a list of string tokens. Finally, we evaluate the LF over the KG to produce a candidate answer.
In the next sections, we describe each step in details.

\subsection{Structured Input computation}
\label{subsec:structured-input}

\paragraph{Hierarchical structure} For each query, we construct the input as a JSON-like structure, consisting of lists of objects with their fields (represented in the left part of Figure \ref{fig:model}). We chose this representation as it allows to incorporate general structured information into the model. A field can be a string, a KG integer ID, a numerical value, or a list thereof.

To construct the input, we start from the last utterances in the dialog: the current query, previous query and previous answer. We first perform NEL to retrieve a list of entities $\mathcal{E}$ (and numerical values) matching the utterances.
The KG is then queried to retrieve additional contextual information: the classes of the entities, and all outgoing and incoming properties from these entities $\mathcal{E}$. This gives a list of properties $\mathcal{P}$. For each property $p \in \mathcal{P}$, we fill several fields: its ID, its name, and an \texttt{Entity IDs} field, which corresponds to all entities $e \in \mathcal{E}$ such that at least one {\em graph operator}  
gives a non-empty result when applied to $e$ and $p$. For instance, in Wikidata, the property \texttt{birth name} (P1477) is filled for Marie Curie but not for Pierre Curie, so the \texttt{Entity IDs} field of the \texttt{birth name} property only contains Marie Curie.

Let us introduce some formal notations, useful to explain the computation of the input's embeddings (Section \ref{subsec:embedding}). The input is a tree where the root corresponds to the whole input, and each leaf contains the primitive values. For a non-leaf node $x$, we denote by $c(x)$ its children.
For instance, in Figure~\ref{fig:model}, the node \texttt{Property 2} has three children (leaves) whose values are \texttt{mother}, \texttt{P25} and \texttt{Q37463}. A node $x$ has also a type, and $T_{\leadsto}(x)$ denotes the types of all nodes on the path from the root to $x$. For instance, for the \texttt{mother} node, $T_{\leadsto}(x)$ is equal to \texttt{(root, property, name)}.
In our setup, the depth of the input is at most $2$.

\paragraph{ID randomization}
Directly giving the entity ID to the model would mean training a categorical classifier with millions of possible outcomes, which would lead to poor generalization. To avoid this, we replace the integer ID with a random one, thereby forcing the model to learn to predict the correct entity from the list of entities in input by copying their randomized entity ID to the output. 

For numerical values, we associate each value to an arbitrary random ID, that the model should learn to copy in the output.
For properties and classes, since there are fewer possibilities in the graph (a few thousand), we do not randomize them. 

\subsection{Embedding}
\label{subsec:embedding}

\paragraph{Preprocessing} We apply BERT tokenization \cite{devlin2018bert} to textual inputs. A vocabulary $\mathcal{V}_t$ is generated for each of the non-textual input types $t$.

\paragraph{Token embedding} The goal of this step is to associate an embedding to each field of each object in the input. We do so by using a learned embedding for each input type: BERT embeddings \cite{devlin2018bert}  for textual inputs, and categorical embeddings for non-textual inputs. When the input is a list (textual tokens or \texttt{Entity IDs} field), we add to this embedding a positional embedding. To reduce the size of the model, list embeddings are averaged into a single embedding.
Formally, the embedding step associates a matrix of embeddings $h(x) \in \mathbb{R}^{1 \times d_h}$ to each leaf of the input tree.

\subsection{Encoding layers}

There are two types of encoding layers: Flattener layers and Transformer layers.

\paragraph{Flattener} The goal of these layers is to compute the embeddings of tree nodes bottom-up. They are successively applied until we are able to compute the embedding of the root node, i.e. of the whole input. This operation can be seen as flattening the JSON-like structure, hence their name. 

Say we want to compute the embedding of some parent node $x$.
An affine projection is first applied to the embedding of each child, then the embedding of the parent node is computed by applying a reduction operation $\mathcal{R}$, which can be either a sum or a concatenation. The weights of the projections are shared between all nodes having the same types $T_{\leadsto}(x)$. For example, all class name nodes - with types \texttt{(root, class, name)} - share the same weights, but they do not share the weights of entity name nodes - with types \texttt{(root, entity, name)}. Hence the embedding of $x$ is
\begin{equation*}
h(x) = \mathcal{R}_{y \in c(x)} \left( \left\{ W_{project}^{T_{\leadsto}(y)} h(y) + b_{project}^{T_{\leadsto}(y)}) \right\} \right).
\end{equation*}
If the reduction is a sum, all children embeddings need to be matrices of the same dimension, and the dimension of the parent embedding is also the same. If the reduction is a concatenation, the dimension of the parent embedding is $\left( \sum_{y \in c(x)} d(y), d_h \right)$.

\paragraph{Transformer} This layer is a classical multi-head Transfomer encoder layer  \cite{vaswaniAttentionAllYou2017a}, taking as input a matrix of embeddings of dimensions $(n, d_h)$, performing self-attention between the input embeddings, and outputting another matrix of the same dimensions. Detailed setup can be found in Appendix \ref{subsec:apx-modeling}.

\paragraph{Architecture}
We apply a first Transformer layer only to the utterances, and in parallel a first Flattener layer with sum reduction to all other inputs. The latter computes one embedding for each object.
We add a positional embedding to each object, to account for its position in the list of objects of the same type. Then we apply a second Flattener layer with concatenation to all outputs of the first layer. This creates a single matrix of embeddings containing the embeddings of all the objects and utterances. Finally, a second Transformer layer is applied to this matrix.

\subsection{Decoding layers}
\label{subsec:decoding}

The output is a list of tokens, which correspond to a prefix representation of the LF tree. Note that the model architecture is grammar agnostic, as this output structure is independent of the grammar and we do not use grammar-guided decoding. The tokens can belong to one of the non textual input types or be a grammar token. Remember that we computed a vocabulary $\mathcal{V}_t$ for all token types $t \in \mathcal{T}$. We augment the vocabulary with a \textsc{stop} token. 

The decoder predicts the output list of tokens iteratively. 
Assume that we are given the first $j$ tokens $y_1, \dots, y_{j}$. We then apply an auto-regressive Transformer decoder on the full sequence, and we condition on the last embedding $h_j$ of the sequence to predict the next token $\hat{y}_{j+1}$. 
Several categorical classifiers are used to predict $\hat{y}_{j+1}$. We first decide whether we should stop decoding:
\begin{equation}
\hat{s}_{j+1} = \argmax p_{stop, j}
\end{equation}
\noindent
where $p_{stop, j} = \softmax(W_{stop} h_{j})$ is a distribution over $\{0, 1\}$ given by a binary classifier. 
If $\hat{s}_{j+1}=1$, the decoding is finished, and we set $\hat{y}_{j+1}$ to \textsc{stop}; otherwise, we predict the type of the token:
\begin{equation}
\hat{t}_{j+1} = \argmax p_{type, j}
\end{equation}
\noindent
where $p_{type, j} = \softmax(W_{type} h_{j})$ is a distribution over $\mathcal{T}$ given by a $|\mathcal{T}|$-class classifier.

Finally, depending on the predicted type, we predict the token itself
\begin{equation}
\hat{y}_{j+1} = \argmax p_{token, j}^{\hat{t}_{j+1}}
\end{equation}
\noindent
where $p_{token, j}^{\hat{t}_{j+1}} = \softmax(W_{token}^{\hat{t}_{j+1}} h_{j})$ is a distribution over $\mathcal{V}_{\hat{t}_{j+1}}$ given by a $\left| \mathcal{V}_{\hat{t}_{j+1}} \right|$-class classifier.

\paragraph{Training} We train by teacher forcing: for some training sample $(\textbf{x}, [y_{1}, \cdots, y_{M}])$ and for each step $j$, the embedding $h_j$ is computed using the expected output at previous steps: $h_j = h(y_{j} ; \textbf{x}, y_{1}, \cdots, y_{j-1})$. The loss is the cross-entropy between the expected output $y_{j+1}$ and the distributions produced by the model. Precisely, let $T: \bigcup_{t \in \mathcal{T}} \mathcal{V}_t \rightarrow \mathcal{T}$ be the mapping which projects tokens to their type. We denote by $p(x)$ the value of a categorical distribution $p$ for the category $x$. Then, omitting the step subscripts $j$, the loss equals
\begin{align*}
l(p, y) = &- \log(p_{stop}(0)) \\
    &- \left[ \log(p_{type}(T(y))) + \log(p_{token}^{T(y)}(y)) \right]
\end{align*}
\noindent
for all steps except the last, and
$l(p, y) = - \log(p_{stop} (1))$ for the last step. $p_{stop}$, $p_{type}$, and $p_{token}$ are computed as explained above. The total loss is obtained by averaging over all training samples and over all steps.

\subsection{Comparison with MaSP architecture \cite{shen-etal-2019-multi}}
\label{subsec:comparison-masp}

Both models follow the semantic parsing approach, where the LF is encoded as a sequence of operators and graph items IDs. Regarding the model input, theirs consist only of the utterances, whereas we add additional KG context structured as a JSON tree. The training method is different: MaSP uses multi-task learning to learn jointly entity linking and semantic parsing, whereas we chain both, and trust the model to pick the good entity. Our approach is simpler in this regard, but we pay this by having a slightly lower performance on Simple Direct questions (see Table \ref{csqa-results}). Finally, we do not use beam search for the decoding, contrarily to them.

\section{Experiments}

\begin{table*}[ht]
\centering
\begin{tabular}{m{0.25\linewidth}m{0.32\linewidth}m{0.32\linewidth}}
\toprule 
& \textbf{CSQA} & \textbf{ConvQuestions} \\
\midrule
Average length of a dialog & 8 turns & 5 turns \\ \hline
Possible change of topic inside a conversation & Yes & No \\ \hline
Answer type & Entities, boolean, quantity & Entities (usually a single one), boolean, date, quantity, string \\ \hline
Entity annotations in the dataset & Yes, with coreference resolution & Only the  \textit{seed entity} (topic of the dialog) and the answer entities \\ \hline
Coreferences in questions & Yes, to the previous turn & Yes, to any preceding turn \\
\bottomrule
\end{tabular}
\caption{\label{dataset-comparison} Some characteristics of the benchmark datasets.}
\end{table*}

\begin{table*}[ht]
\centering
\begin{tabular}{l|S[table-format=3]|S[table-format=2.2]S[table-format=2.2]S[table-format=2.2]S[table-format=2.2]S[table-format=2.2]S[table-format=2.2]}
\toprule
\textbf{Methods} &  & \textbf{D2A} & \textbf{D2A+ES} & \textbf{S2A+MAML} & \textbf{MaSP} & \textbf{OAT (Ours)}   \\
\midrule
\textbf{Question type} & \textbf{\# Examples} & & & {\textbf{F1}} & & \\
\hline
Simple (Direct) & 82\si{\kilo} & 91.41 & 83.00 & \bfseries 92.66 & 85.18 & 82.69 \\
Simple (Coreferenced) & 55\si{\kilo} & 69.83 & 64.62 & 71.18 & 76.47 & \bfseries 79.23 \\
Simple (Ellipsis) & 10\si{\kilo} & 81.98 & 83.94 & 82.21 & 83.73 & \bfseries 84.44 \\
Logical & 22\si{\kilo} & 43.62 & 72.93 & 44.34 & 69.04 & \bfseries 81.57  \\
Quantitative & 9\si{\kilo} & 50.25 & 63.95 & 50.30 & 73.75 & \bfseries 74.83 \\
Comparative & 15\si{\kilo} & 44.20 & 55.05 & 48.13 & 68.90 & \bfseries 70.76  \\
\hline
\textbf{Question type} & \textbf{\# Examples}  & & & {\textbf{Accuracy}} & & \\ 
\hline
Verification (Boolean) & 27\si{\kilo} & 45.05 & 45.80 & 50.16 & 60.63 & \bfseries 66.39 \\
Quantitative (Count) & 23\si{\kilo} & 40.94 & 41.35 & 46.43 & 43.39 & \bfseries 71.79 \\
Comparative (Count) & 15\si{\kilo} & 17.78 & 20.93 & 18.91 & 22.26 & \bfseries 36.00 \\
\hline
\hline
\textbf{Total Average} & 260\si{\kilo} & 64.47 & 64.75 & 66.54 & 70.56 & \bfseries 75.57 \\
\bottomrule
\end{tabular}
\caption{\label{csqa-results} QA performance on CSQA. The metric is the F1 score for question types above the vertical separator, and accuracy for those under. The Total Average score is an average over all question types.}
\end{table*}

Additional comments about the datasets, setups, and additional results can be found in the appendix.

\subsection{Datasets}

We use two weakly supervised conversational QA datasets to evaluate our method, Complex Sequential Question Answering (CSQA)\footnote{\url{https://amritasaha1812.github.io/CSQA}} \cite{sahaComplexSequentialQuestion2018} and ConvQuestions\footnote{\url{https://convex.mpi-inf.mpg.de}} \cite{christmannLookYouHop2019}, both grounded in Wikidata\footnote{\url{https://www.wikidata.org}}. CSQA consists of about $1.6$M turns in $200$k conversations ($152$k/$16$k/$28$k splits), versus $56$k turns in $11$k conversations ($7$k/$2$k/$2$k splits) for ConvQuestions.

CSQA was created by asking crowd-workers to write turns following some predefined patterns, then turns were stitched together into conversations. The questions are organized in different categories, e.g. simple, logical or comparative questions.

For ConvQuestions, crowd-workers wrote a $5$-turn dialog  in a predefined domain (\textit{e.g.} Books or Movies). The dialogs are more realistic than in CSQA, however at the cost of a smaller dataset.

As presented in Table~\ref{dataset-comparison}, the datasets have different characteristics, which make them an interesting test bed to assess the generality of our approach.

\subsection{CSQA Experimental Setup}
\label{subsec:csqa-setup}

\paragraph{Metrics} To evaluate our grammar, we report the coverage, i.e. the percentage of training questions for which we found a candidate Logical Form.

To evaluate the QA capabilities, we use the same metrics as in \citet{sahaComplexSequentialQuestion2018}. F1 Score is used for questions whose answers are entities, while accuracy is used for questions whose answer is boolean or numerical. 
We don't report results for ``Clarification'' questions, as this question type can be accurately modeled with a simple classification task, as reported in Appendix \ref{subsec:clarification}. Similarly the average metric ``Overall'' (as defined in \citealt{sahaComplexSequentialQuestion2018}) is not reported in Table~\ref{csqa-results}, as it depends on ``Clarification'', but can be found in the Appendix.

\paragraph{Baselines} We compare our results with several baselines introduced in Section~\ref{sec:related-work}: D2A \cite{guoDialogtoActionConversationalQuestion2018}, D2A+ES \cite{shenEffectiveSearchLogical2019}, S2A+MAML \cite{guo-etal-2019-coupling}, and MaSP \cite{shen-etal-2019-multi}. 

\subsection{ConvQuestions Experimental Setup}

\paragraph{Metrics} We use the coverage as above, and the P@1 metric as defined in \citet{christmannLookYouHop2019}.

\paragraph{Baseline} The only baseline to our knowledge is \textsc{Convex} \cite{christmannLookYouHop2019}, which casts the problem to a node classification task. For comparison, we tried to make our setup as close as possible to theirs, and refer to Appendix \ref{subsec:comparison-node-classif} for details.

\paragraph{Data augmentation} 
Given the small size of the dataset, we merge it with two other data sources: CSQA, and $3.6$M examples generated by random sampling. The latter are single-turn dialogs made from graph triplets, \textit{e.g.} the triplet  \textit{(Marie Curie, instance of, human)} generates the dialog: \textit{Q: Marie Curie instance of? A: Human}. More details are given in Appendix \ref{subsec:random-generation}.
The Conv\-Questions dataset is upsampled to match the other data sources sizes.

\subsection{Named Entity Linking setup}
\label{subsec:nel}
We tried to use a similar setup as baselines for fair comparison. For CSQA, the previous gold answer is given to the model in an oracle-like mode, as in baselines. In addition, we use simple string matching between utterances and names of Wikidata entities to retrieve candidates that are given in input to the model. For ConvQuestions, we use the gold seed entity (as in the \textsc{Convex} baseline we compare with), and the Google Cloud NLP service. We refer to Appendices \ref{appendix:silver-lf} and \ref{subsec:apx-modeling} for details. 

Regarding CARTON~\cite{plepi2021context}, their results are not directly comparable as their model uses gold entity annotations as input and hence is not affected by NEL errors. This different NEL setup does have a strong influence on the performance, as running our model on CSQA with a setup similar to CARTON improves our Total Average score by over 10\%. We refer to Appendix \ref{subsec:oracle-setup} for details. More generally, a more thorough study of the impact of the NEL step on the end-to-end performance would be an interesting direction of future work (see also Section \ref{subsec:error-analysis}).

\subsection{Results}
\label{subsec:results}

\paragraph{Our grammar reaches high coverage.} With approximately the same numbers of operators as in baselines, we improve the CSQA coverage by $16\%$, as presented in Table \ref{table:csqa-coverage-per-question-type}. The improvement is particularly important for the most complex questions.  We reach a coverage of $86.2\%$ on ConvQuestions, whose questions are more varied than in CSQA.

\begin{table}
\centering
\begin{tabular}{lS[table-format=2]S[table-format=2]S[table-format=3.1]}
\toprule
\textbf{Question type} & \textbf{D2A} & \textbf{D2A+ES} & \textbf{Ours} \\
\midrule
Comparative  & 28.6   & 45 & \bfseries 84.9 \\
Logical      & 48.2  & 92 & \bfseries 100.0 \\
Quantitative & 58.1  & 62 & \bfseries 91.1 \\
Simple       & 94.4  & 96 & \bfseries 99.7 \\
Verification & 77.9  & 85 & \bfseries 91.4 \\
\hline
\hline
Overall      & 74.3  & 80 & \bfseries 96.2 \\
\bottomrule
\end{tabular}
\caption{\label{table:csqa-coverage-per-question-type} Coverage per question type for CSQA. 
}
\end{table}

Most queries can be expressed as relatively shallow LFs in our grammar, as illustrated by Table~\ref{table:lf-depth}. This is especially interesting for the ConvQuestions dataset, composed of more realistic dialogs. On CSQA, the average depth of our LFs (2.9) is slightly slower than with D2A grammar (3.2). 

\begin{table}
\centering
\begin{tabular}{lS[table-format=1.1]S[table-format=1.1]S[table-format=1.1]S[table-format=1.1]}
\toprule
\textbf{Depth} & {\textbf{1}} & {\textbf{2}} & {\textbf{3}} & {\textbf{4+}} \\
\midrule
CSQA (D2A) & 0.0  & 47.0 & 30.9 & 22.0 \\
CSQA (Ours)                    & 5.5  & 67.9 & 7.4 & 19.2 \\
ConvQuestions              & 53.9 & 43.7 & 2.4 & 0.0 \\
\bottomrule
\end{tabular}
\caption{\label{table:lf-depth} Silver LF depth distribution for both datasets.}
\end{table}

\paragraph{We improve the QA performance over baseline on both datasets.}
For CSQA, our model outperforms baselines for all question types but Direct Simple questions, as shown in Table~\ref{csqa-results}. 
Overall, our model improves the performance by $5\%$. 
For Conv\-Questions, Table~\ref{convquestions-results} shows that our model improves over the baseline for all domains but one, yielding an overall improvement of $4.7\%$. A precise evaluation of the impact of the various components of our KG-QA approach (grammar, entity linking, model inputs, model architecture, size of the training data, etc.) on the end-to-end performance was out of the scope of this paper, and is left for future work. Nevertheless, the fact that we are able to improve over baselines for two types of Simple questions and for Logical questions, for which the grammar does not matter so much, as these question types correspond to relatively shallow LFs, suggests that our proposed model architecture is effective.

\begin{table}
\centering
\begin{tabular}{l|S[table-format=2.1]|S[table-format=2.1]S[table-format=2.1]}
\toprule
\textbf{Domain} & \textbf{$1^{\text{st}}$ turn} & \textbf{Follow-up} & \textbf{\textsc{Convex}} \\ \midrule
Books   & 68.1 & \bfseries 20.9 & 19.8 \\
Movies  & 54.2 & \bfseries 31.3 & 25.9 \\
Music   & 37.5 & 18.1 & \bfseries 19.0 \\
Soccer  & 43.8 & \bfseries 22.8 & 18.8 \\
TV      & 66.3 & \bfseries 31.8 & 17.8 \\
\hline
\hline
Overall & 54.0 & \bfseries 25.0 & 20.3 \\
\bottomrule
\end{tabular}
\caption{\label{convquestions-results} ConvQuestions results by domain. The first two columns are our results. The baseline (Oracle+\textsc{Convex}) only reports follow-up turns.} 
\end{table}

\subsection{Error analysis}
\label{subsec:error-analysis}

\paragraph{CSQA} By comparing the silver and the predicted LFs on $10\si{\kilo}$ random errors, we could split the errors in two main categories: first, the LF general form could be off, meaning that the model did not pick up the user intent. Or the form of the LF could be right, but (at least) one of the tokens is wrong. Table \ref{table:csqa-error-analysis} details the error statistics.
The most frequent errors concern entity disambiguation.
There are two types of errors: either the correct entity was not part of the model input, due to insufficient recall of the NEL system. Or the model picked the wrong entity from the input due to insufficient precision. 
It is known that the noise from NEL strongly affects model performance \citep{shen-etal-2019-multi}. We tried an oracle experiment with perfect recall NEL (see Appendix \ref{subsec:oracle-setup}), which corroborates this observation, in particular for Simple questions. As we focused on modeling complex questions, improving NEL was not our main focus, but would an interesting direction for future work, in particular via multi-task approaches \cite{shen-etal-2019-multi}.

\begin{table}[ht]
\centering
\begin{tabular}{lS[table-format=2.2]S[table-format=2.2]}
\toprule
\textbf{Error category} & \textbf{Overall} & \textbf{Simple Dir.}   \\
\midrule
LF general form & 31.8 & 24.1 \\
Entity ID token & 36.2 & 38.2 \\
\;\;\;\; insuff. recall & 17.1 & 16.8 \\
\;\;\;\; insuff. precision & 19.6 & 21.6 \\
Property ID token & 4.2 & 2.9 \\
Class ID token & 24.7 & 37.6 \\
Grammar token & 11.6 & 2.6 \\
\bottomrule
\end{tabular}
\caption{\label{table:csqa-error-analysis} Distribution of errors in CSQA. The numbers are (non exclusive) percentages.
 We also report statistics for the Simple Direct type, as it is the largest.}
\end{table}

\paragraph{ConvQuestions} We manually analyzed 100 examples. Errors were mostly due to the LF general form, then to a wrong property token.

\paragraph{The model learns the grammar rules.} In all inspected cases, the predicted LF is a valid LF according to the grammar, i.e. it could be evaluated successfully. This shows that grammar-guided decoding is not needed to achieve high performance.

\section{Conclusion}

For the problem of weakly-supervised conversational KG-QA, we proposed Object-Aware Transformer, a model capable of processing structured input in a JSON-like format. This allows to flexibly provide the model with structured KG contextual information. We also introduced a KG grammar with increased coverage, which can hence be used to model a wider range of queries.
These two contributions are fairly independent : on the one hand, since the model predicts LFs as a list of tokens, it is grammar agnostic, and thus it could be used with another grammar. On the other hand, the grammar is not tied to the model, and can be used to generate training data for other model architectures.
Experiments on two datasets validate our approach.
We plan to extend our model to include a richer KG context, as we believe there is significant headroom for improvements.


\section*{Ethical considerations}

This work is not connected to any specific real-world application, and solely makes use of publicly available data (KG and QA datasets). The predominant ethical concern of the paper is the computing power associated with the experiments. To limit the energy impact of the project, we did not perform hyper-parameter tuning for model training. Using the formulas from \citet{patterson2021carbon}, we estimate the GHG emissions associated with one run of model training to be approximately 29 - 42 kg of CO2e. For the silver LF generation, we iterated on a small subset of the datasets, then computed the LFs for the entire  datasets as a one-off task.

\section*{Acknowledgments}

Authors thank Yasemin Altun, Eloïse Berthier, Guillaume Dalle, Maxime Godin, Clément Mantoux, Massimo Nicosia, Slav Petrov, as well as the anonymous reviewers, for their constructive feedback, useful comments and suggestions. Authors also thank Daya Guo for gracefully providing results computed with the D2A grammar. P. Marion was supported by a stipend from Corps des Mines.

\bibliographystyle{acl_natbib}
\bibliography{anthology,acl2021}




\appendix

\section{Clarification Questions in CSQA}
\label{subsec:clarification}

Take the following dialog as example:

\vspace{0.3cm}

\noindent
\begin{tabular}{m{0.1\linewidth}m{0.78\linewidth}}
\hline
\multirow[c]{2}{\linewidth}{T1} & \em Can you tell me which cities border Verderio Inferiore? \\
 & Cornate d’Adda, Bernareggio, Robbiate \\
 \hline
\multirow[c]{2}{\linewidth}{T2} & \em And which cities flank that one? \\
 & Did you mean Robbiate? \\
 \hline
\multirow[c]{2}{\linewidth}{T3} & \em No, I meant Cornate d’Adda. \\
 & Bottanuco, Busnago, Trezzo sull’Adda \\
 \hline
\end{tabular}

\vspace{0.3cm}

The second turn is a ``Clarification'' question: the system asks the user for disambiguation. The disambiguation question usually takes the form ``Did you mean'', followed by an entity chosen among the previous turn answers. This choice appears to be entirely random. For this reason, we found that it would not be very interesting to try to predict this entity, as baselines propose. Hence we only ask the model to predict that the question is a Clarification (via a special \texttt{clarification} operator). 

We report in Table \ref{csqa-extended-results} the scores for Clarification questions, as well as the ``Overall'' score, as defined in \citet{sahaComplexSequentialQuestion2018}.
The results are not directly comparable as the baseline systems report an F1 score, while our approach uses accuracy.

\begin{table*}[ht]
\centering
\begin{tabular}{l|S[table-format=3]|S[table-format=2.2]S[table-format=2.2]S[table-format=2.2]S[table-format=2.2]S[table-format=2.2]S[table-format=2.2]}
\toprule
\textbf{Question type} & \textbf{\# Examples} & \textbf{D2A} & \textbf{D2A+ES} & \textbf{S2A+MAML} & \textbf{MaSP} & \textbf{OAT (Ours)}   \\
\midrule
\hline
Clarification & 12\si{\kilo} & 18.31 & 36.66 & 19.12 & 80.79 & \bfseries 99.63 \\
\hline
\hline
\textbf{Overall} & 206\si{\kilo} & 62.88 & 72.02 & N/R & 79.26 & \bfseries 81.49 \\
\bottomrule
\end{tabular}
\caption{\label{csqa-extended-results} QA performance on CSQA, including ``Clarification'' questions. The ``Overall'' metric is the average F1 scores of the following question types: ``Simple (Direct)'', ``Simple (Coreferenced)'', ``Simple (Ellipsis)'', ``Logical'', ``Quantitative'', ``Comparative'' and ``Clarification''.}
\end{table*}

\section{Detailed experimental setup}
\label{sec:detailed-setup}

\subsection{Meta-operators}
\label{subsec:meta-operators}

Take the example given in the main paper: 
\emph{``Which musical instrument is played by the maximum number of persons?''}. 
The corresponding LF is:

{
\small
\noindent \texttt{argmax( \\
\indent cardinality( \\
\indent \indent follow\_property( \\
\indent \indent \indent for\_each( \\
\indent \indent \indent \indent members(\textit{musical instrument})), \\
\indent \indent \indent \textit{played by}))) \\
}}

Assume that the KG contains exactly two musical instruments, piano and violin, i.e. \texttt{members(\textit{musical instrument})} equals \texttt{\{\textit{piano, violin}\}}.

\texttt{for\_each} creates a dictionary of entities. Each (key, value) pair corresponds to one entity in the argument of \texttt{for\_each}, where the key is the entity itself and the value is a singleton set containing the entity. Here \texttt{for\_each(\{\textit{piano, violin}\})} gives the following dictionary:

{
\small
\noindent \texttt{\{ \\
\indent piano:  \{piano\}, \\
\indent violin: \{violin\} \\
\}}
}

We then apply the same computation to each of the dictionary values, while keeping the keys untouched.
In our example, we apply the expression

{
\small
\noindent \texttt{cardinality( \\
\indent follow\_backward(., \textit{played by}) \\
)}, 
}

\noindent which gives the result

{
\small
\noindent \texttt{\{piano: 20392, violin: 7918\}}.
}

Finally, an aggregation operator is computed over the values, and the result is a subset of the keys. In the example, \texttt{argmax} returns the set of keys associated with the maximum values, here \texttt{\{piano\}}. In other cases, we want to return all the keys associated to a non-empty value, \texttt{arg} allows to do so.

\subsection{Silver LF generation}
\label{appendix:silver-lf}

\paragraph{Wikidata version} For CSQA, we used the preprocessed version of Wikidata made available by the authors, which contains $21.2$M triplets over $12.8$M entities and $567$ distinct properties. For ConvQuestions, we used a more recent version of Wikidata, containing $1.1$B triplets over $91.8$M entities and $7869$ distinct properties.

\paragraph{Named Entity Linking} For ConvQuestions, we use gold entity annotations and Google Cloud NLP entity linking service. For CSQA, we use gold entity annotations.

To resolve the coreferences, in ConvQuestions, we use entity annotations from previous utterances during the silver LF generation step. In CSQA, since coreferences are already resolved by the gold annotations, we just use annotations from the current utterance.

\paragraph{Simplifying the BFS} We observed that reaching a depth of $4+$ is needed for some queries (see Table~\ref{table:lf-depth} of the main paper), but is impractical by exhaustive BFS, as the size of the space of LFs grows very quickly with their depth.  To improve the efficiency, we used the following ideas:

\begin{itemize}
\item \textbf{Stopping criteria to abort the exploration:} timeout $t_{\max}$ and maximum depth $d_{\max}$.
\item \textbf{Type checking:} by leveraging the operators' signatures (presented in Table~\ref{list-operators} of the main paper), we only construct legal LFs.
\item \textbf{Putting constraints on the form of the LF:} we manually forbid certain combinations of operators, e.g. \texttt{follow\_backward} after \texttt{follow\_property}.
\item \textbf{Restriction of the list of operators:} for Conv\-Questions, we use the graph operators, the numerical operators, \texttt{is\_in}, and \texttt{get\_first}. The removal of some set operators and of meta-operators strongly reduces the complexity of the BFS. For CSQA, all operators are needed, but we add more constraints in order to keep the BFS simple enough.
\end{itemize}

We choose $d_{\max}=3$ for ConvQuestions and $d_{\max}=7$ for CSQA, and $t_{\max}=1200$ seconds. 

All LFs found by BFS are evaluated over the KG, which gives candidate answers. We keep the LFs whose candidate answers have the highest F1 score w.r.t. the gold answer. The minimal F1 score for keeping a LF is $0.3$. 

\paragraph{Scores for LF ranking} The BFS often returns several LFs (with the top F1 score, as explained above), among which some are spurious: they do not correspond to the semantic meaning of the question, but their evaluation over the KG yields the correct result by chance. As we keep only one for training, we need a way to rank the candidate LFs. We use the following heuristic scores to do so:
\begin{itemize}
\item \textbf{Complexity:} the score is $1-\nicefrac{(d-1)}{(d_{\max}-1)}$ where $d$ is the depth of the LF and $d_{\max}$ is defined above.
\item \textbf{Property lexical matching:} for each property appearing in the LF, we compute the Jaccard index of the words appearing in its name and of the words of the question. 
\item \textbf{Annotation coverage:} among the entities retrieved by NEL, we compute the percentage of entities which appear in the LF.
\end{itemize}

As these three scores are between $0$ and $1$, we average them and keep the LF with the highest total score. We found that this simple method is a good way to reduce spurious LFs, which are often either too complex or not matching lexically the question.

\subsection{Random examples generation}
\label{subsec:random-generation}

To generate the random examples used for data augmentation for ConvQuestions training, we first sample uniformly $80$k entities from the graph. Then, for each entity, we generate a conversation for each triplet that links it to other entities. The question text is made by stitching the  entity name and the property name. For instance, the triplet \texttt{(Marie Curie, native language, Polish)} generates the dialog: \textit{Q: Marie Curie native language? A: Polish}. We also generate variants where the property name is replaced by aliases, which are alternative names given in Wikidata. For instance, the property \texttt{native language} has aliases \textit{first language}, \textit{mother tongue}, \textit{language native} and \textit{L1 speaker of}. When the question or answer has more than 256 characters, we eliminate the example.

\subsection{Modeling}
\label{subsec:apx-modeling}

\paragraph{Wikidata version} As in \ref{appendix:silver-lf}.

\paragraph{Named Entity Linking} NEL is performed again, this time to create the structured context in input to the model. Due to the randomization step described in Section~\ref{subsec:structured-input}, missing entities in the input cannot be retrieved by the model, so we want to have a high NEL recall. The trade-off is that the NEL precision is low, meaning that we have many spurious entities in the input, which the model has to learn to ignore. 

For CSQA, we use simple string matching between the utterances and the names of Wikidata entities. Note that in our model, as well as in all CSQA baselines (in particular \citealp{guoDialogtoActionConversationalQuestion2018, shen-etal-2019-multi}), the previous gold answer is given as input to the model in an oracle-like setup.

For ConvQuestions, we use the gold seed entity and the Google Cloud NLP entity linking service.

To resolve coreferences, we use entities from the dialog history: all preceding turns for ConvQuestions and only the previous turn for CSQA.

\paragraph{Implementation details} 
We tokenize the input using the BERT-uncased tokenizer. All embeddings in the model have dimension $768$. The two transformer encoders share the same configuration: $2$ layers each, with output dimension $768$, intermediate dimension $2048$, $12$ attention heads, and dropout probability set to $0.1$. The model has $260$M parameters. The transformer implementation is based on publicly available BERT code. We initialize the word embedding from a BERT checkpoint, but do \emph{not} load the transformer layer weights, instead training them from scratch. We train for $600$k steps with batch size $128$, using the ADAM optimizer \cite{kingmaAdamMethodStochastic2017} with learning rate \num{3e-5}. Training takes around $14$ hours on 16 TPU v3 with $32$ cores.

\section{Comparison with baselines}

\subsection{Comparison with D2A grammar}
\label{subsec:comparison-d2a}

\begin{table*}[htbp]
\centering
\begin{tabular}{m{0.23\linewidth}m{0.40\linewidth}m{0.26\linewidth}}
\toprule 
\textbf{Intent} & \textbf{Question example} & \textbf{Missing in D2A} \\
\midrule
Textual form reasoning & What was Elvis Presley given name? & string type \\
\hline
Numerical reasoning & What actor plays the younger child? & \texttt{get\_value}, \texttt{for\_each}, \texttt{argmin} \\
\hline
Numerical reasoning & How old is the younger child? & \texttt{min} \\
\hline
Selection of the members of a class & Which television programs have been dubbed by at least 20 people ? & \texttt{members}  \\
 \hline
Temporal reasoning & What is the number of seasons until 2018? & \texttt{for\_each}, \texttt{get\_value}, \texttt{lesser\_than}, \texttt{arg} \\ 
\hline
Ordinal reasoning & What was the first episode date? & \texttt{get\_first} \\
\bottomrule
\end{tabular}
\caption{\label{table:d2a-comparison} Examples of questions that are difficult to model with the D2A grammar. Examples are mostly chosen from ConvQuestions, as their questions look more realistic than CSQA.}
\end{table*}

The D2A \citep{guoDialogtoActionConversationalQuestion2018} grammar is the main baseline in previous KG-QA works.
We compare with the grammar implemented in their open-sourced code\footnote{\tiny{\url{https://github.com/guoday/Dialog-to-Action/blob/bb2cbb9de474c0633bac6d01c10eca24c79b951f/BFS/parser.py}}}, which is a bit different from the published one. Numbers for D2A in Tables \ref{table:csqa-coverage-per-question-type} and \ref{table:lf-depth} were computed thanks to the results of the BFS gracefully provided by the authors.

Table \ref{table:d2a-comparison} presents some intents which we are able to model in our grammar and are not straightforward to model with D2A grammar. First,  \textit{textual form reasoning} corresponds to questions about string attributes of entities, which are not included in the D2A grammar. Second, to handle \textit{numerical and temporal reasoning}, computations based on numerical values are needed, which is not possible with the D2A grammar. Finally, the D2A grammar does not model the order of relations in the graph and the selection of class members, which we start to tackle with respectively the \texttt{get\_first} and \texttt{members} operators.

\subsection{Comparison with node classification approaches}
\label{subsec:comparison-node-classif}

An alternative to the semantic parsing approach is to train classifiers to predict entities as nodes of the KG. A precise comparison of both approaches is out of the scope of this paper. Nevertheless, we think that the semantic parsing approach is better suited to our purpose of modeling complex questions. For instance, complex intents involving numerical comparisons can be expressed naturally by a LF, but would be difficult to perform using solely node classifiers. Examples include the numerical and temporal reasoning in Table \ref{table:d2a-comparison}. Additional examples include \textit{The series consists of which amount of books?} (ConvQuestions) or \textit{Which television programs have been dubbed by at least 20 people ?, How many episodes is it longer than the second longest season out of the three?} (CSQA).

\textsc{Convex} \cite{christmannLookYouHop2019} is an example of such an approach. It is an unsupervised graph exploration method: at each turn, a subgraph is expanded by matching the utterance with neighboring entities. Then a candidate answer is found in the subgraph by a node classifier. 
On our side, we propose a semantic parsing approach that makes use of entities annotated by an external entity linking service.
This is a similar setup to the CSQA baselines \cite{shen-etal-2019-multi}, which we re-purposed for ConvQuestions in order to assess the quality of our proposal on another dataset. 
In order to be closer to the \textsc{Convex} baseline, we changed our CSQA setup by applying the entity linker only to the questions' text and not to the answers' text.
In addition, as we use the gold seed entity, we compare with the Oracle+\textsc{Convex} setup of \citet{christmannLookYouHop2019}, which also uses the gold seed entity (and the gold first turn answer entity). Finally, we make use of data augmentation to train our model on ConvQuestions, whereas the baseline does not.

\subsection{BERT or no BERT, that is the question}
The baselines of Table \ref{csqa-results} do not use BERT.
MaSP authors provide an additional BERT variant of their model that uses a fine-tuned BERT base architecture. The Total Average score of this variant is 72.60\%, which is 2\% above their vanilla variant and 3\% under our model. Since we are only using the word embeddings (loaded from a publicly available BERT base checkpoint) and not loading the transformer layer weights, we decided to compare with the vanilla variant of MaSP, and not the BERT one. 
Finally, CARTON is using a pre-trained BERT base model as a sentence encoder.

\section{Additional results}

\subsection{Coverage results}

We present in Table~\ref{table:convquestions-coverage-per-domain} the coverage per domain for ConvQuestions. Besides, the evolution of the coverage over turns is stable for both datasets, hence we do not report this result. 

\begin{table}
\centering
\begin{tabular}{lS[table-format=2.1]}
\toprule
\textbf{Domain} & \textbf{Coverage} \\
\midrule
Books   & 88.6 \\
Movies  & 87.6 \\
Music   & 90.0 \\
Soccer  & 78.6 \\
TV      & 86.2 \\
\hline
Overall & 86.2 \\
\bottomrule
\end{tabular}
\caption{\label{table:convquestions-coverage-per-domain} ConvQuestions coverage per domain.}
\end{table}

\subsection{Performance over turns}

Tables \ref{csqa-over-turns} and \ref{convquestions-over-turns} show the evolution of the performance over turns for both datasets. For CSQA, the performance drops after the first two turns, then remains constant. For ConvQuestions, the performance decreases throughout the turns. There is a sharp decrease after the first turn, probably because it is simpler as there is no coreference or ellipsis. The different behavior between the datasets may be due to the realism of ConvQuestions.

\begin{table}[ht]
\centering
\begin{tabular}{cS[table-format=2]S[table-format=2]S[table-format=2]S[table-format=2]S[table-format=2]}
\toprule
\textbf{Turns} & {\textbf{0}} & {\textbf{1}} & {\textbf{2}} & {\textbf{3}} & {\textbf{4}} \\
\midrule
\textbf{Score} & 85 & 87 & 75 & 74 & 74 \\
\\
\toprule
\textbf{Turns} & {\textbf{5-6}} & {\textbf{7-8}} & {\textbf{9-10}} & {\textbf{11-12}} & {\textbf{13+}} \\
\midrule
\textbf{Score} & 75 & 75 & 75 & 74 & 74 \\
\end{tabular}
\caption{\label{csqa-over-turns} Average performance over turns for CSQA. For brevity, we average over turn ranges after turn 5.}
\end{table}

\begin{table}[ht]
\centering
\begin{tabular}{cS[table-format=2]S[table-format=2]S[table-format=2]S[table-format=2]S[table-format=2]}
\toprule
\textbf{Turn} & {\textbf{0}} & {\textbf{1}} & {\textbf{2}} & {\textbf{3}} & {\textbf{4}} \\
\midrule
\textbf{Av. P@1} & 54 & 35 & 20 & 29 & 15 \\
\bottomrule
\end{tabular}
\caption{\label{convquestions-over-turns} Performance over turns for ConvQuestions.}
\end{table}

\subsection{Oracle setup and comparison with CARTON}
\label{subsec:oracle-setup}

CARTON \cite{plepi2021context} gives the entities annotated in the dataset as part of the model input (entities appearing in the previous turn and in the current question), contrarily to the models in Table~\ref{csqa-results} which all use an entity linker. For a fair comparison, we tested our model in an oracle setup, where we also give the gold annotations as input. As shown in Table \ref{table:csqa-results-oracle}, the Total Average score of our model increases by $10\%$ w.r.t. the baseline approach. The improvement is particularly important for the most simple question types (Simple and Logical Questions). In this setup, our performance is $8\%$ higher than CARTON, and we obtain a better score for 7 out of 10 question types.

\begin{table}[ht]
\centering
\begin{tabular}{lS[table-format=2.2]S[table-format=2.2]}
\toprule
\textbf{Question type} & \textbf{CARTON} & \textbf{Ours}   \\
\midrule
Simple (Direct) & 85.92 & \bfseries 96.95 \\
Simple (Coreferenced) & 87.09 & \bfseries 94.77 \\
Simple (Ellipsis) & 85.07 & \bfseries 96.66 \\
Logical & 80.80 & \bfseries 95.54  \\
Quantitative & \bfseries 80.62 & 76.44 \\
Comparative & 62.00 & \bfseries 76.66  \\
Verification (Boolean) & \bfseries 77.82 & 67.02 \\
Quantitative (Count) & 57.04 & \bfseries 75.89 \\
Comparative (Count) & \bfseries 38.31 & 35.10 \\
\hline
\hline
\textbf{Total Average} & 77.89 & \bfseries 85.85 \\
\bottomrule
\end{tabular}
\caption{\label{table:csqa-results-oracle} QA performance on CSQA in oracle mode.}
\end{table}

\subsection{Further error analysis}

An alternative approach for error analysis is to assess the performance of the decoding classifiers (see Section \ref{subsec:decoding}) in a teacher forcing setup, \textit{i.e.} to assess how often they predict the next token correctly, given the true previous tokens. Table~\ref{table:lf-token-accuracy} reports the results on the eval split of both datasets. 
The results corrobate the analysis presented in Section \ref{subsec:error-analysis}. First, the model learns the grammar rules, as it nearly always predicts the good token type. For CSQA, the most frequent errors concern entity ID and class ID. For ConvQuestions, they concern primarily entities and properties.

\begin{table}
\centering
\begin{tabular}{lS[table-format=2.2]S[table-format=2.2]}
\toprule
\textbf{Metric} & \textbf{CSQA} & \textbf{ConvQuestions} \\
\midrule
Token type      & 99.88 & 94.77 \\
\hline
Grammar token   & 98.86 & 82.12 \\
Entity ID       & 92.47 & 50.79 \\
Property ID     & 99.45 & 30.26 \\
Class ID        & 94.64 & N/A \\
Numerical value & 99.91 & N/A \\
\hline
Avg. token      & 97.70 & 61.53 \\
\bottomrule
\end{tabular}
\caption{\label{table:lf-token-accuracy} LF token accuracy metrics, on the eval splits. 
For Conv\-Questions, Class ID, numerical value and their relative operators are not used (see \ref{appendix:silver-lf}).}
\end{table}

\subsection{Case study}

Table \ref{table:kg-context-properties} presents examples from ConvQuestions where we are able to predict the good LFs, although there exists very similar properties in the graph. The textual forms of the questions are not sufficient to infer the good property to use, implying that the model had to learn elements from the graph structure in order to answer correctly these questions. Nevertheless, Table \ref{table:lf-token-accuracy} shows that there is still significant room for improvement in that direction.

\begin{table}
\centering
\begin{tabular}{m{0.64\linewidth}m{0.27\linewidth}}
\toprule 
\textbf{Question} & \textbf{Property} \\
\midrule
When did Seinfeld first air? & start time (P580) \\
When did Camp Rock come out? & publication date (P577) \\
\hline
Who screen wrote it? & screenwriter (P58) \\
Who wrote it? & author (P50) \\
\hline
What country are they from? & country (P17) \\
Belleville of which country? & country (P17) \\
What country did the band Black Sabbath originally come from? & country of origin (P945) \\
What country is Son Heung-min from originally? & country for sport (P1532) \\
\bottomrule
\end{tabular}
\caption{\label{table:kg-context-properties} Examples of ConvQuestions questions for which the model was able to pick up the good property, although there are very similar properties in the graph.}
\end{table}

\end{document}